\documentclass[sigconf]{acmart}

\renewcommand\footnotetextcopyrightpermission[1]{} 
\usepackage{color,array}
\usepackage{graphicx}
\usepackage{float} 
\usepackage{hyperref}       
\usepackage{url}            
\usepackage{booktabs}       
\usepackage{amsfonts}       
\usepackage{nicefrac}       
\usepackage{microtype}      
\usepackage{lipsum}
\usepackage{fancyhdr}       
\usepackage{graphicx}       
\usepackage{subcaption}
\usepackage{amsmath}

\newcommand{\METHOD}{FaceX}
\AtBeginDocument{%
  \providecommand\BibTeX{{%
    \normalfont B\kern-0.5em{\scshape i\kern-0.25em b}\kern-0.8em\TeX}}}

\begin{document}
\pagestyle{plain} 

\title{\METHOD: Understanding Face Attribute Classifiers through Summary Model Explanations}

\author{Ioannis Sarridis$^{1,2}$ ~~~~ Christos Koutlis$^{1}$ ~~~~ Symeon Papadopoulos$^1$   ~~~~
Christos Diou$^2$ \vspace{3pt} \\ 
$^1$Information Technologies Institute, CERTH, Thessaloniki, Greece\\ 
$^2$Department of Informatics and Telematics, Harokopio University, Athens, Greece\\ 
{\tt\small \{gsarridis, ckoutlis, papadop\}@iti.gr} ~~~~ {\tt\small \{isarridis, cdiou\}@hua.gr} 
}

\renewcommand{\shortauthors}{Ioannis Sarridis, Christos Koutlis, Symeon Papadopoulos, and Christos Diou}

\begin{abstract}
EXplainable Artificial Intelligence (XAI) approaches are widely applied for identifying fairness issues in Artificial Intelligence (AI) systems.
However, in the context of facial analysis, existing XAI approaches, such as pixel attribution methods, offer explanations for individual images, posing challenges in assessing the overall behavior of a model, which would require labor-intensive manual inspection of a very large number of instances and leaving to the human the task of drawing a general impression of the model behavior from the individual outputs. 
Addressing this limitation, we introduce \METHOD, the first method that provides a comprehensive understanding of face attribute classifiers through \textit{summary model explanations}. 
Specifically, \METHOD\ leverages the presence of distinct regions across all facial images 
to compute a region-level aggregation of model activations,
allowing for the visualization of the model's region attribution across 19 predefined regions of interest in facial images, such as hair, ears, or skin.
Beyond spatial explanations, \METHOD\ enhances interpretability by visualizing specific image patches with the highest impact on the model's decisions for each facial region within a test benchmark. 
Through extensive evaluation in various experimental setups, including scenarios with or without intentional biases and mitigation efforts on four benchmarks, namely CelebA, FairFace, CelebAMask-HQ, and Racial Faces in the Wild, \METHOD\ demonstrates high effectiveness in identifying the models' biases.
\end{abstract}

\begin{teaserfigure}
    \centering
    \includegraphics[width=\linewidth]{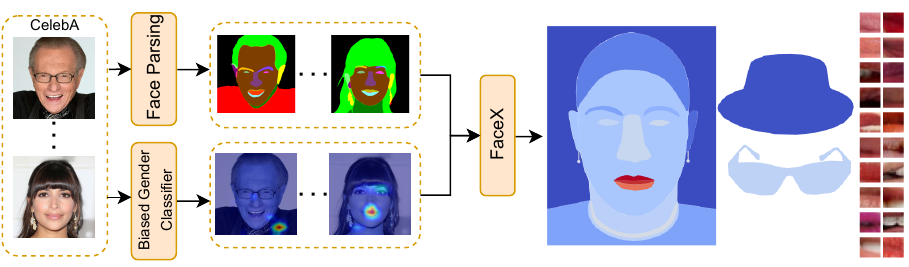}
    \caption{
    \METHOD\ employs 19 facial regions and accessories to provide explanations (left: face regions, right: hat and glasses). Blue to red colors indicate low to high importance, respectively.  The provided illustration answers the questions ``where does a model focus on?'' and ``what visual features trigger its focus?'' through heatmap and high-impact patches visualizations, respectively. This example depicts a biased \texttt{gender} classifier trained on CelebA that effectively uses the \texttt{Wearing\_Lipstick} attribute as a shortcut to predict \texttt{Gender}. Note that \METHOD\ is compatible with any face dataset.}
    \label{fig:facex}
\end{teaserfigure}
\maketitle

\section{Introduction}

In the quickly evolving landscape of Artificial Intelligence (AI), facial image processing models have seen unprecedented integration into applications affecting billions of users worldwide. From unlocking smartphones \cite{taigman2014deepface} to 
emotion recognition technologies \cite{zhao2014exploring}, computer vision based face analysis has become an integral part of our daily lives. Although such advancements have ushered in a new era of possibilities, they also raise several ethical and societal concerns. 

\begin{figure}[t]
    \centering
    \begin{subfigure}{0.32\linewidth}
        \centering
        \includegraphics[width=\linewidth]{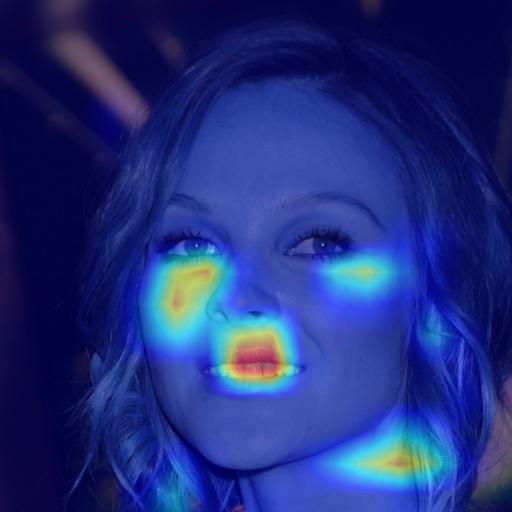}
    \end{subfigure}
    \begin{subfigure}{0.32\linewidth}
        \centering
        \includegraphics[width=\linewidth]{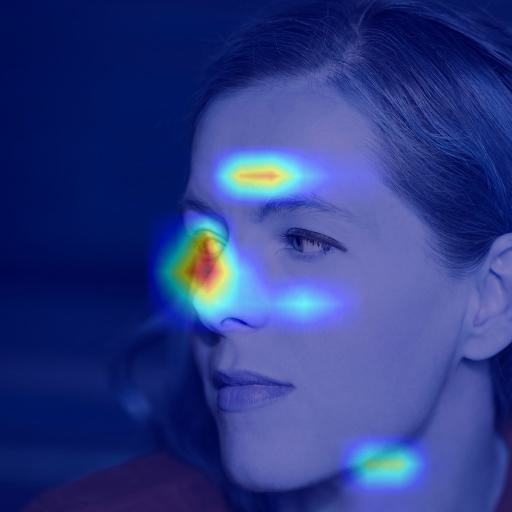}
    \end{subfigure}
    \begin{subfigure}{0.32\linewidth}
        \centering
        \includegraphics[width=\linewidth]{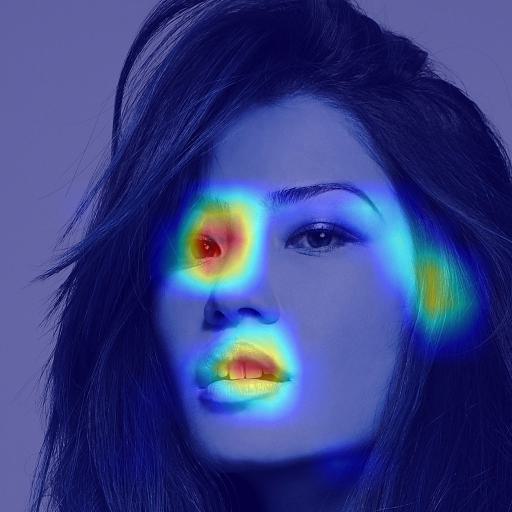}
    \end{subfigure}
    
    \begin{subfigure}{0.32\linewidth}
        \centering
        \includegraphics[width=\linewidth]{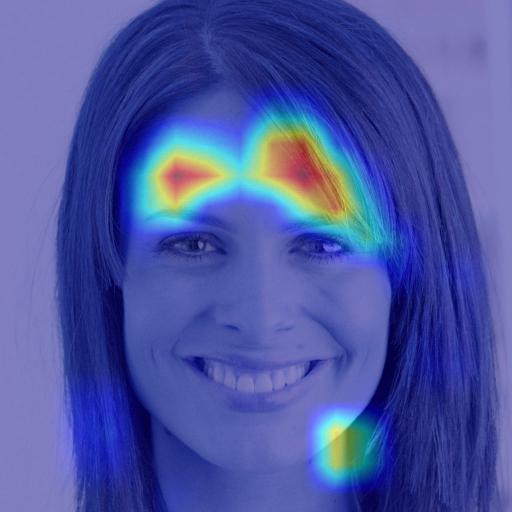}
    \end{subfigure}
    \begin{subfigure}{0.32\linewidth}
        \centering
        \includegraphics[width=\linewidth]{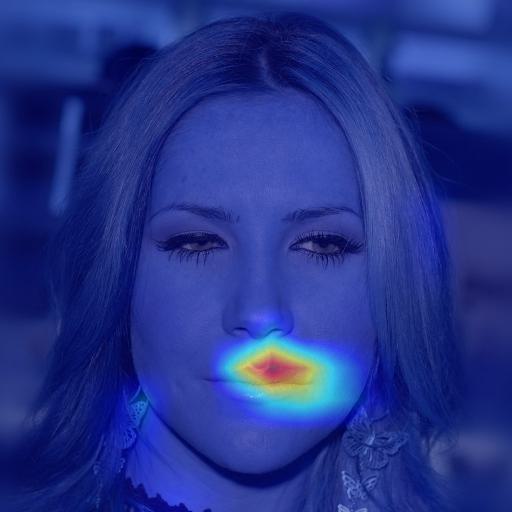}
    \end{subfigure}
    \begin{subfigure}{0.32\linewidth}
        \centering
        \includegraphics[width=\linewidth]{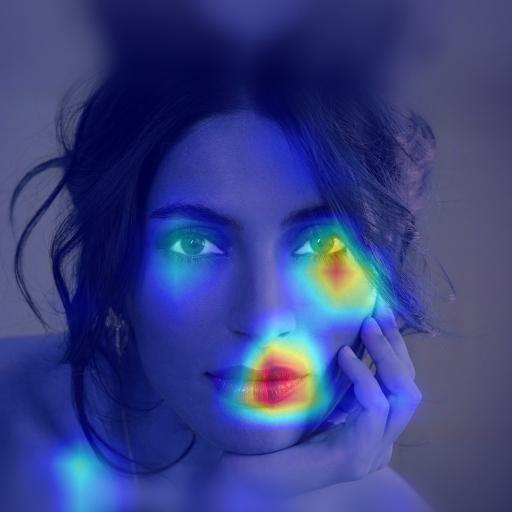}
    \end{subfigure}
    
    \caption{Grad-CAM instance-level explanations for six random samples for a gender classifier biased towards the \texttt{Wearing\_Lipstick} attribute. Two key limitations are evident: a) there is an inconsistent attribution on the mouth region, hindering the user's ability to pinpoint where the biased attribute occurs; b) interpreting the visual characteristics of the region of interest (i.e., lipstick) is not straightforward. The corresponding \METHOD\ summary model explanation is provided in Figure~\ref{fig:facex}.}
    \label{fig:indiv_fail}
\end{figure}

AI bias has been among the key concerns in relation to 
the wide deployment of facial AI systems in high-stakes contexts,  ranging from influencing law enforcement decisions \cite{flores2016false} to shaping hiring processes \cite{pena2020bias} and impacting identity verification procedures \cite{sarridis2023towards,melzi2024frcsynIF,melzi2024frcsyn}, which could disproportionately affect certain population groups.
In an effort to delve deeper into the reasons of AI bias, Explainable Artificial Intelligence (XAI) methods \cite{adadi2018peeking,schultze2023explaining,karayil2019focus} aims at shedding light on the decision making mechanisms of AI models. However, in the realm of computer vision models, existing XAI methods, especially those providing visual explanations in the form of pixel attribution maps, with Grad-CAM \cite{selvaraju2017grad} being the most popular, are limited to individual explanations for model decisions. This poses challenges in assessing the overall behavior of a model. Specifically, while individual explanations can indeed point to biased model decisions in given examples, they cannot help with more general statements about the behavior of an AI model, as this would require a labor-intensive manual inspection of many images in a test set, and then the synthesis of the individual outputs 
into a general assessment of model behavior. This manual process is not objective, repeatable, or feasible in many cases.
This limitation is prominent in the context of facial analysis, where the correlation between the target attribute (e.g., \texttt{gender}, \texttt{race}, and \texttt{age}) and certain other attributes within the training data can lead to biased models \cite{sarridis2023flac}. For instance, CelebA \cite{liu2015deep}, a widely used dataset for facial analysis tasks, exhibits high correlation between \texttt{gender} and several attributes, such as \texttt{Blond\_Hair}, \texttt{Wearing\_Earrings}, and \texttt{Wearing\_Lipstick}  \cite{zhang2018examining}. Such correlations are exploited by facial classifiers and, consequently, the resulting models heavily rely on these proxy attributes (i.e., ``shortcuts'') - which are easier to learn than the actual target - to make their decisions \cite{zhang2018examining}. In such cases, XAI methodologies that rely on instance-level explanations cannot accurately depict the overall behavior of a model. Figure~\ref{fig:indiv_fail} illustrates an example of a biased \texttt{gender} classifier (i.e., it focuses on the \texttt{Wearing\_Lipstick} attribute) 
to highlight the limitations of instance-level explanations: solely inspecting a few test instances cannot adequately capture the overall region attribution or the visual features influencing the model's decisions.

To bridge this gap, we present \METHOD , a methodology designed to provide a comprehensive understanding of model decisions in facial analysis. \METHOD\ aims at summarizing the behavior of a model with respect to 19 predefined regions of interest in facial images, such as hair, ears, and skin \cite{farl}, and additionally provides appearance-oriented insights through high-impact image patches, drawn from test set samples, aiming to not only highlight where the model focuses but also  show which visual appearances trigger high model activations.

Specifically, \METHOD\ computes a region-level aggregation of model instance-level attributions, summarizing the model's output with respect to each region of interest. Then, spatial explanations, offered through a heatmap visualization over an abstract face prototype, provide in-depth understanding of the weight of each facial region (or accessory)  on the model decision. Additionally, \METHOD\ visualizes the high-impact image patches for each region, revealing not only where the model focuses but also helping the human analyst understand why certain features are influential (see Figure~\ref{fig:facex}). 
This dual approach of spatial explanation (understanding where the model focuses) and appearance-oriented insights (understanding the impact of specific image patches) sets \METHOD\ apart as a powerful tool for identifying biases in facial analysis systems and acts as a comprehensive lens, allowing practitioners, researchers, and developers to scrutinize the entire spectrum of model behavior. 
In extensive evaluation, \METHOD\ is tested on controlled scenarios with several combinations of targets and biased attributes, achieving high precision in discovering both single and multi-attribute biases. 
Furthermore, real scenarios with known biases are considered - without intentionally injecting bias - where \METHOD\ successfully uncovers the biases introduced by data.

In summary, the paper makes the following contributions:
(i) Introduces \METHOD , a methodology that, for the first time, provides summary model explanations for face attribute classifiers, (ii) Provides both facial region attribution and insights into the specific features influencing the model's focus by visualizing high-impact image patches within activated regions, (iii) Provides analysis for controlled single-attribute bias scenarios involving four datasets, namely CelebA \cite{liu2015deep}, FairFace \cite{karkkainen2019fairface}, CelebAMask-HQ \cite{lee2020maskgan}, and Racial Faces in the Wild (RFW) \cite{wang2019racial}, and (iv) Provides a comprehensive analysis involving controlled multi-attribute bias scenarios, experiments on real use cases, and comparison between models before and after applying a bias mitigation approach. The extensive evaluation provides a thorough understanding of \METHOD 's effectiveness in discovering biases of different type in various scenarios. Code is available at \url{https://github.com/gsarridis/faceX}.
\looseness=-1
\section{Related Work}
\textbf{Bias in facial analysis.} Numerous works highlight the bias existence in facial analysis models and data \cite{khalil2020investigating, sarridis2023towards,fabbrizzi2022survey,joo2020gender}. A study presented in \cite{yucer2022measuring} suggests using facial phenotype attributes to discover and assess racial bias in face recognition tasks. 
Also, biases in gender classifiers are investigated in \cite{ngan2015face}, revealing that middle-aged males are more likely to be correctly classified than other age or gender groups. An in-depth analysis of intersectional biases across gender, age, and race in face recognition models is presented in \cite{sarridis2023towards}, suggesting that models discriminate against certain population groups, with Asian females suffering from the most severe bias. Similarly, an analysis presented in \cite{krishnan2020understanding} highlights the impact of imbalanced training data on gender classifiers across race-gender intersections.
Furthermore, \cite{taati2019algorithmic} explores biases against clinical populations in landmark detection and expression recognition models. The findings suggest that these models exhibit lower accuracy when applied to older individuals with dementia.
A study in \cite{buolamwini2018gender} revealed significant disparities in classification performance between darker-skinned females and lighter-skinned males in commercial gender classifiers.
Similarly, the impact of skin tone and gender in facial expression detection is investigated in \cite{deuschel2020uncovering}.

\textbf{Explainable Artificial Intelligence.}
In the landscape of XAI, numerous methodologies have emerged to illuminate the intricate behavior of complex machine learning models \cite{liu2016towards,ribeiro2018anchors,lundberg2017unified,selvaraju2017grad, kocielnik2019will,lim2019these}.
In considering explanations for computer vision models, approaches like Grad-CAM \cite{selvaraju2017grad}, Grad-CAM++ \cite{chattopadhay2018grad}, HiRes-CAM \cite{draelos2020use}, 
Layer-CAM \cite{jiang2021layercam}, Score-CAM \cite{wang2020score}, SmoothGrad \cite{smilkov2017smoothgrad}, LIME \cite{ribeiro2016should}, Layer-wise Relevance Propagation (LRP) \cite{montavon2019layer}, and Axiomatic Attribution \cite{sundararajan2017axiomatic} offer diverse visual formats as a means of explaining the outputs of a model. However, all of these approaches pertain to individual explanations, which constitutes a significant limitation when it comes to the overall behavior assessment of a facial analysis model. Our proposed \METHOD\ aims to build on top of such individual explanations to synthesize summary model explanations.
\looseness=-1

The inherent limitations of instance-level approaches have motivated several works to build on them for generating summary model explanations. While such methods are often referred to as ``global'' in the literature, we prefer the term ``summary'' model explanations 
as the distinction between ``global'' and ``local'' is ambiguous for methods offering model explanations through multiple instance-level explanations.
SP-LIME \cite{ribeiro2016should} reduces numerous instance-level attributions into a concise set by selecting the most important of them. Similarly, Concept Relevance Propagation (CRP) \cite{achtibat2023attribution} visualizes only a few reference samples for each class. In the same way, Spray \cite{lapuschkin2019unmasking}, ACE \cite{ghorbani2019towards}, and GAM \cite{ibrahim2019global} leverage clustering techniques to provide a summary of instance-level attributions. The common drawback of these methods is their inability to offer a single comprehensive explanation that encapsulates all instance-level insights. Instead, they yield multiple instance-level explanations in structures that aid user interpretation of the model's overall behavior (e.g., clusters).
Capitalizing on this geometrical consistency, \METHOD\ addresses the aforementioned limitations and offers for the first time a single summary model explanation through facial region attribution. It should be highlighted that \METHOD\ can be employed on any domain demonstrating a similar geometrical abstraction (e.g., fine-grained vehicle classification).

\section{Methodology}
\label{sec:method}
\subsection{Problem formulation}
The problem of providing explanations pertaining to the model's region attribution can be formulated as follows.
Let $(\mathbf{X}_i, y_i, \mathbf{t}_i)$ be the $i$-th sample of the dataset $\mathcal{D}$, where $\mathbf{X}_i\in \mathbb{R}^{H\times W\times 3}$ is an input color image with height $H$ and width $W$, $y_i\in\mathcal{Y}$ the target label, and $\mathbf{t}_i \subseteq \mathcal{T}$ a set of attributes potentially introducing bias.
Then, the classification model is denoted as $f(\cdot)$ and its predictions as $\hat{y}_i = f(\mathbf{X}_i)$. To provide summary model explanations we need to map instance-level attributions onto certain facial regions, so let us also define the region masks for the $i$-th sample, $\mathbf{M}_{ir}\in \{0,1\}^{H\times W}, r\in (0,1,\dots, R)$ where $R$ is the number of the facial regions of interest (e.g., hair) and $\mathbf{M}_{ir}$ is a binary mask, where a pixel is assigned the value $1$ if it belongs to the region of interest ($0$ otherwise). Collectively, all region masks for the $i$-th sample are denoted as $\mathbf{M}_{i}$.
Additionally, the normalized attribution values provided by an XAI approach, such as Grad-CAM, are denoted as $\mathbf{G}_i\in [0,1]^{H\times W}$. Having defined $\mathbf{M}_i$ and $\mathbf{G}_i$, the target is to exploit this information to map the pixel attribution to certain regions and thus, allow for generating a summary map (i.e., for all the test samples) involving all the $R$ facial regions of interest. Complementary to region attribution, visual features attribution is provided through visualizing the image patches with the highest impact on the model's decision for each facial region. 
Let $Z$ denote the length of a square patch, then we partition uniformly $\mathbf{X}_i$ into patches of size $Z \times Z$, i.e., $\mathbf{P}_{i,q}\in \mathbb{R}^{Z\times Z\times 3},q\in\{1,2,\ldots\,Q\}$, where $Q$ is the number of patches.
The set of patches for $i$-th image can be defined as $\mathcal{P}_i = \{\mathbf{P}_{i,1}, \mathbf{P}_{i,2}, \ldots, \mathbf{P}_{i,Q}\}$. 
Then, the target is to derive the set of top-$k$ patches for each region $r$, which is defined as $\mathcal{S}_{r} = \{\mathbf{S}_{r}^1, \mathbf{S}_{r}^2, \ldots, \mathbf{S}_{r}^k\}$, where $\mathbf{S}_{r}^j$ represents the $j$-th highest activated patch in region $r$ across all test samples.

\subsection{Face Parsing}
\label{sec:farl}
\METHOD\  relies on the facial region masks, $\mathbf{M}_i$, to produce explanations for facial analysis models. The task of generating these masks is termed ``face parsing'' and can be considered as a segmentation problem. To the best of our knowledge, among the datasets used for face parsing, the CelebAMask-HQ \cite{lee2020maskgan} offers the richest annotations in terms of the number $R$, of facial regions (i.e., skin, left/right brow, left/right eye, eyeglasses, left/right ear, earrings, nose, mouth, upper/lower lip, neck, necklace, cloth, hair, hat, and background). Thus, without loss of generality, we define facial regions in alignment with this dataset. However, note that \METHOD\ is compatible with any other face parsing protocol (i.e., set of regions).
In instances where CelebAMask-HQ serves as the test benchmark for \METHOD , with the desired targets $\mathcal{Y}$ and protected attributes $\mathcal{T}$ inherent to CelebAMask-HQ, we set the masks $\mathbf{M}_i$ equal to the ground truth masks provided by the dataset. In the more general case, however, where \METHOD\ is applied to arbitrary test face images, a face parsing model becomes essential for predicting $\mathbf{M}_i$. To this end, we opted for utilizing the Facial Representation Learning (FaRL) \cite{farl} approach that offers powerful pre-trained transformer backbones 
for several face analysis tasks, including face parsing. 
\subsection{Individual explanations}
After obtaining masks $\mathbf{M}_i$, the next task is to derive individual explanations for each sample in $\mathcal{D}$. This process can be accomplished by employing any local XAI method that generates activations in the form of a heatmap as output. In the context of this paper, we focus on CNN model architectures, and thus we utilize Grad-CAM, a widely applied and robust approach. The Grad-CAM method computes the gradient of the score for a particular class with respect to the feature maps of a convolutional layer. This gradient information is then used to produce a weighted combination of the feature maps, highlighting regions in the input image that strongly influence the model's decision for the given class.

In particular, to compute the Grad-CAM score $g_y(\mathbf{X}_i)$ for a class $y\in \mathcal{Y}$ and an input image $\mathbf{X}_i$, first the gradient of the classification layer's output score for class $y$, $l^y$, w.r.t. the feature map activations $A^k$ ($k$ denotes the channel index) is calculated and then, global average pooling is applied to these gradients to derive the neuron importance weights: 
\begin{equation}
    a_k^y = \frac{1}{Z}\sum_i\sum_j \frac{\partial l^y}{\partial A_{i,j}^k}
\end{equation}
The final score is obtained by applying a ReLU activation to the weighted combination of feature maps:
\begin{equation}
g_y(\mathbf{X}_i) = \text{ReLU}\left(\sum_{k} \alpha_{k}^y \cdot A^k\right)    
\end{equation}
This score represents the regions in the input image that contribute the most to the model's decision for a specific class. 

Notably, there is a direct connection between the resolution of the input image and the resolution of the Grad-CAM output. As the input image progresses through convolutional layers of a CNN, undergoing successive convolutions and pooling operations, the spatial dimensions of the feature map gradually decrease. Consequently, the resulting Grad-CAM heatmap is directly influenced by the reduced spatial resolution of the feature map. 
This interdependence highlights the need to consider the initial resolution of the input image when interpreting and analyzing the Grad-CAM visualizations, as finer details may be affected by the downscaling process inherent in the convolutional neural network architecture. Therefore, we suggest resizing the test samples to an adequate size (e.g. 512x512), even if the model is trained with images of lower dimensions, in cases where the model architecture allows for it (e.g., ResNets).
\looseness=-1
\subsection{Summary model explanations}
After obtaining facial region masks $\mathbf{M}_i$ and instance-level activations values $\mathbf{G}_i$ for each sample in the dataset $\mathcal{D}$, we leverage these components to produce comprehensive explanations. The goal is to assess how much the model's attribution aligns with the $R$ facial regions with respect to a certain class.

Particularly, we compute the Hadamard product between the instance-level scores and the corresponding region mask for each region $r\in \{1,2, \cdots, R\}$ and divide by the total number of active pixels within the region mask. This measure can be referred to as Intersection over Region (IoR): 

\begin{equation}
  \text{IoR}_{i,r} = \frac{\sum_{h, w}\left(\mathbf{G}_i \odot \mathbf{M}_{i,r}\right)_{h,w}}{\sum_{h,w}\left(\mathbf{M}_{i,r}\right)_{h,w}} 
\end{equation}
where operator $\odot$ denotes the Hadamard product and the sum is over all image pixels ($h=1,\dotsc,H$, $w=1,\dotsc,W$). 
In other words, $\text{IoR}_{i,r}$ measures the focus within region $r$ for the $i$-th sample, as a percentage of the maximum possible focus for that region. Then, the average $\text{IoR}_{r}$ across all relevant (i.e., samples involving $r$) test samples represents the overall attribution of region $r$: 
\begin{equation}
    \text{IoR}_{r} = \frac{\sum_{i\in\mathcal{D}} \text{IoR}_{i,r}}{N_r}
\end{equation}
where $N_r$ denotes the total number of masks pertaining to region $r$. It should be noted that dividing by $N_r$ is important, considering that not all regions of interest may be present in every test sample (e.g., some samples may depict individuals without earrings, eyeglasses, etc.). Collectively, the IoR values are defined as: 
\begin{equation}
    \text{IoR} = \{\text{IoR}_{1}, \text{IoR}_{2}, \dots, \text{IoR}_{R} \}
\end{equation}

The obtained normalized IoR values allow \METHOD\ to provide a comprehensive understanding of the model's focus on different facial regions with respect to certain decisions, e.g., positive class. To facilitate a more intuitive understanding of these values and to highlight the differences between regions, we have created a face prototype,
which is an abstract face partitioned in 19 facial regions. By visualizing these IoR values in the form of a heatmap on this standardized face prototype, \METHOD\ provides an intuitive means to compare and interpret region attribution for an attribute classifier.

\subsection{High-impact image patches}
While IoR values offer valuable insights into where the model directs its attention, they inherently lack information regarding the specific features within these regions that contribute to a model's decisions. In other words, IoR provides the ``where'' but not the ``what'' of the model's focus.
For example, IoR can effectively highlight the hair region for a gender classifier biased toward hair color, but it cannot provide insights into the visual characteristics within that region that affect the model's decisions (e.g., blond hair).

To address this limitation and enhance the interpretability of the model's decisions, we propose the visualization of high-impact patches within the activated regions. 
Each patch, $\mathbf{P}_{i,q}$, can also be represented by a binary mask indicating its position in the image, i.e., $\mathbf{P}'_{i,q}\in \{0,1\}^{H\times W},q\in\{1,2,\ldots\,Q\}$. Then, the set of patch masks are defined as $\mathcal{P}'_i = \{\mathbf{P}'_{i,1}, \mathbf{P}'_{i,2}, \ldots, \mathbf{P}'_{i,Q}\}$. Next, we leverage the extracted instance-level scores, denoted as $\mathbf{G}_i$, to measure the attribution within these patches concerning specific facial regions:

\begin{equation}
\label{eq:patches}
    V_{i,q,r} = \sum_{h,w} \left(\mathbf{G}_{i} \odot \mathbf{M}_{i,r} \odot \mathbf{P}'_{i,q}\right)_{h,w}
\end{equation}
with the $k$-highest values across all samples for region $r$ defining the set $\mathcal{V}_{max_r}$.
Then, the most influential patches for region $r$ are defined as follows:
\begin{equation}
    \mathcal{S}_{r} = \{\mathbf{P}_{i,q}|V_{i,q,r}\in \mathcal{V}_{max_r}\}
\end{equation}
This patch-based approach allows us to pinpoint the specific image patches contributing most significantly to the model's decisions within each facial region.

\section{Experimental Setup}
Evaluation of \METHOD\ involves a set of experiments to assess its performance in detecting biases present in a model. We curated subsets of the CelebA dataset that exhibit high correlations between target attributes (\texttt{Gender} and \texttt{Age}) and specific facial features, introducing a deliberate bias into the data. The overarching aim is to evaluate \METHOD 's ability to identify these biases, reflected in high IoR values for pertinent facial regions.
Beyond the controlled experiments that explore correlations between the target and individual attributes, we consider evaluation to scenarios involving multiple attribute correlations. This expanded scope enables us to assess \METHOD 's performance in more complex bias scenarios, where the interplay of multiple attributes contributes to the model's predictions.
Moreover, evaluation involves experiments conducted on established benchmarks to showcase the practical utility of \METHOD\ in real-world scenarios, illustrating its effectiveness in unveiling model biases beyond the confines of controlled experiments.
\subsection{Datasets}
For the evaluation of the proposed approach, we employ four datasets, namely CelebA \cite{liu2015deep}, CelebAMask-HQ \cite{lee2020maskgan}, FairFace \cite{karkkainen2019fairface}, and RFW \cite{wang2019racial}. 
\looseness=-1

\textbf{Training datasets.} CelebA and FairFace are used for training a variety of attribute classifiers. CelebA consists of more than 200,000 facial images annotated with 40 binary attributes, we consider \texttt{Gender} and \texttt{Age} as the target attributes. For the controlled experiments, we inject bias by selecting a subset of images such that there is a 99\% correlation between the target attribute and a certain facial attribute. For example, in the case of \texttt{Gender} as the target and \texttt{Wearing\_Lipstick} as the correlated attribute, 99\% of females wear lipstick, while 99\% of males do not.  The FairFace dataset consists of 108,000 facial images and is designed to be balanced in terms of \texttt{Gender} and \texttt{Race}. Here, \texttt{Gender} is the target and \texttt{Race} is the protected attribute. For the controlled experiments on  FairFace, we force a 99\% correlation between \texttt{Gender} and \texttt{Race}.

\textbf{Test benchmarks.} CelebAMask-HQ and RFW are used as test datasets. In particular,  CelebAMask-HQ is a subset of CelebA, consisting of 30,000 images with 19 facial region annotations. Note that the samples belonging to both CelebA and CelebAMask-HQ are removed from the training data.  Finally, the RFW dataset is a test benchmark consisting of 40,000 facial images and it equally represents the different races. 

\subsection{Implementation details and evaluation protocol}
For controlled experiments, we artificially inject bias by selecting a dataset subset that demonstrates 99\% co-occurrence between the target class and a facial attribute. In particular, we considered facial attributes that are directly connected with certain facial regions, i.e., Blond\_Hair, Eyeglasses, Smiling, Wearing\_Earrings, Wearing\_Lipstick, Wearing\_Necklace, and Wearing\_Hat, to facilitate the evaluation. All models involved in the evaluation are trained using the Adam optimizer with an initial learning rate of 0.001 that decays by a factor of 0.1 at 1/3 and 2/3 of the total training epochs, the weight decay is equal to $10^{-4}$ and the batch size is set to 128. Models are trained for 20 epochs in total. For the controlled experiments, we test them on a balanced (fair) subset of the test dataset with respect to the target and the attribute introducing the bias. As evaluation metric, we use the ranking position, which refers to the position of the target region in the \METHOD\ output ranking. Note that this performance metric operates under the assumption that biased models should consistently prioritize the biased region over other regions, a premise that might not align with the behavior of deep learning models given their inherent characteristics. 
In other words, when training a deep learning model with data containing a spurious correlation in a specific region, there is no guarantee that the model will exclusively focus on this region to make decisions. However, due to the lack of a solid ground truth, we make this assumption to enable a quantitative evaluation of \METHOD, rather than relying solely on qualitative results.
All the experiments were conducted on a single NVIDIA RTX-3090 GPU.
\begin{figure*}[t]
    \centering
        \begin{subfigure}{0.16\linewidth}
        \centering
        \includegraphics[width=\linewidth]{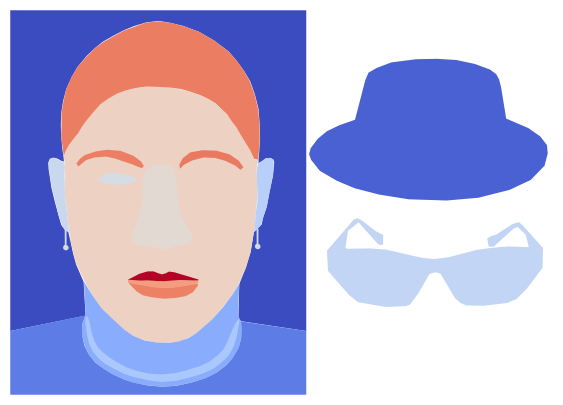}
        \caption{\footnotesize{Gender - Blond\_Hair}} 
    \end{subfigure}
    \begin{subfigure}{0.16\linewidth}
        \centering
        \includegraphics[width=\linewidth]{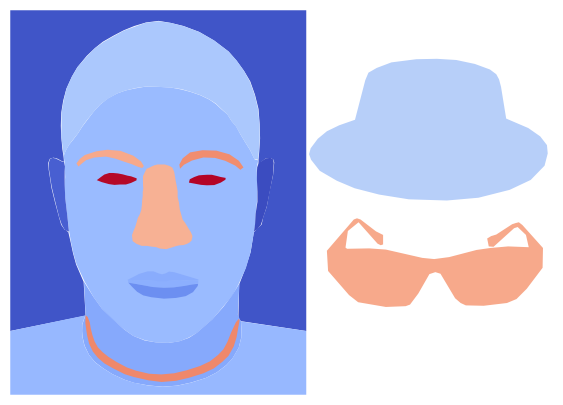}
        \caption{\footnotesize{Gender - Eyeglasses}}
    \end{subfigure}
    \begin{subfigure}{0.16\linewidth}
        \centering
        \includegraphics[width=\linewidth]{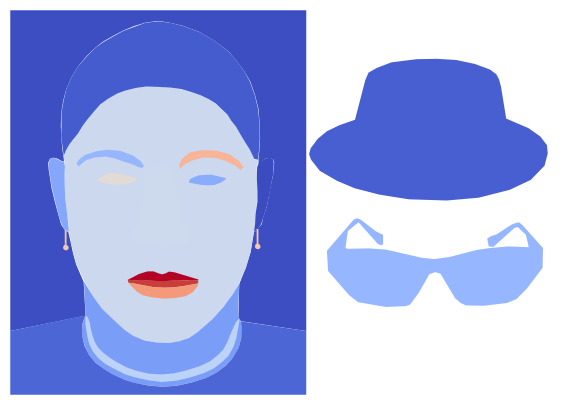}
        \caption{\footnotesize{Gender - Earrings}}  
    \end{subfigure}
        \begin{subfigure}{0.16\linewidth}
        \centering
        \includegraphics[width=\linewidth]{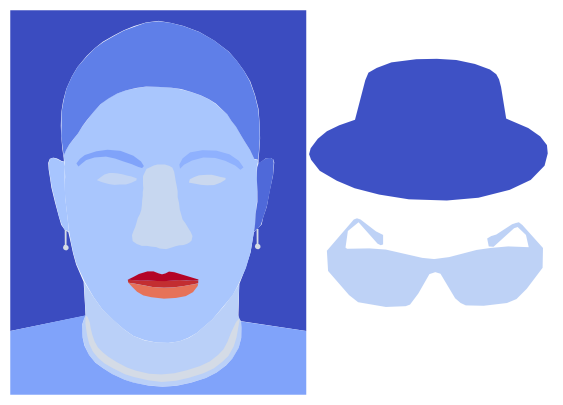}
        \caption{\footnotesize{Gender - Lipstick}} 
    \end{subfigure}
    \begin{subfigure}{0.16\linewidth}
        \centering
        \includegraphics[width=\linewidth]{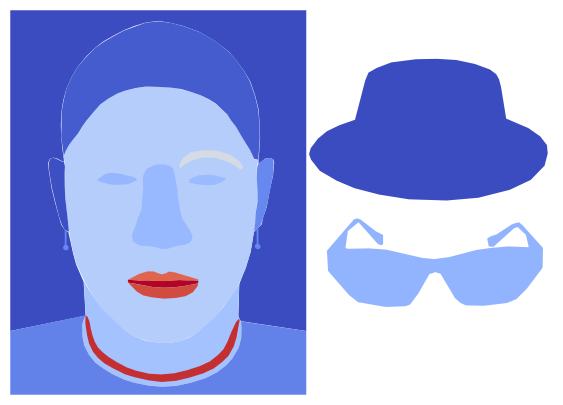}
        \caption{\footnotesize{Gender - Necklace}}
    \end{subfigure}
    \begin{subfigure}{0.16\linewidth}
        \centering
        \includegraphics[width=\linewidth]{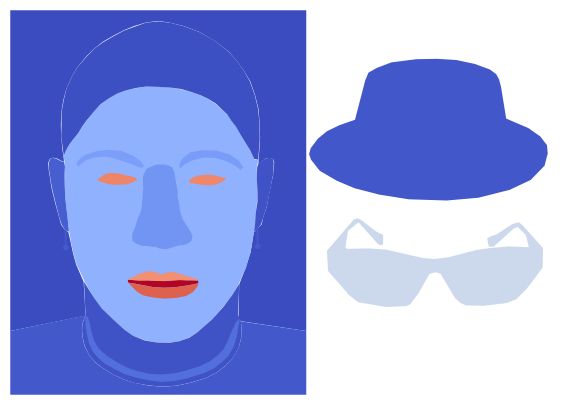}
        \caption{\footnotesize{Gender - Smiling}}
    \end{subfigure}
    
    \begin{subfigure}{0.16\linewidth}
        \centering
        \includegraphics[width=\linewidth]{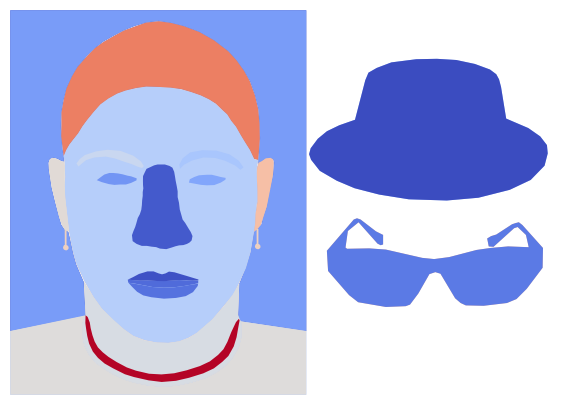}
        \caption{\footnotesize{Age - Blond\_Hair}}
    \end{subfigure}
    \begin{subfigure}{0.16\linewidth}
        \centering
        \includegraphics[width=\linewidth]{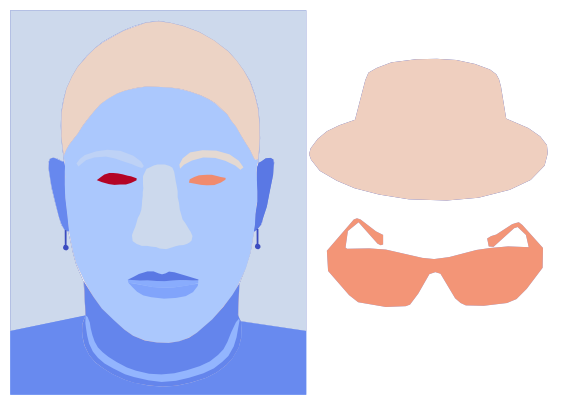}
        \caption{\footnotesize{Age - Eyeglasses}}
    \end{subfigure}
    \begin{subfigure}{0.16\linewidth}
        \centering
        \includegraphics[width=\linewidth]{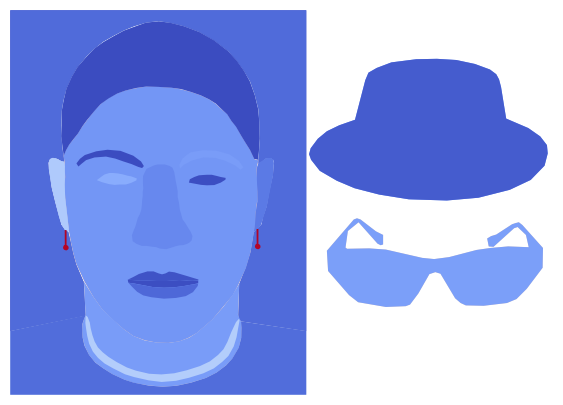}
        \caption{\footnotesize{Age - Earrings}}
    \end{subfigure}
        \begin{subfigure}{0.16\linewidth}
        \centering
        \includegraphics[width=\linewidth]{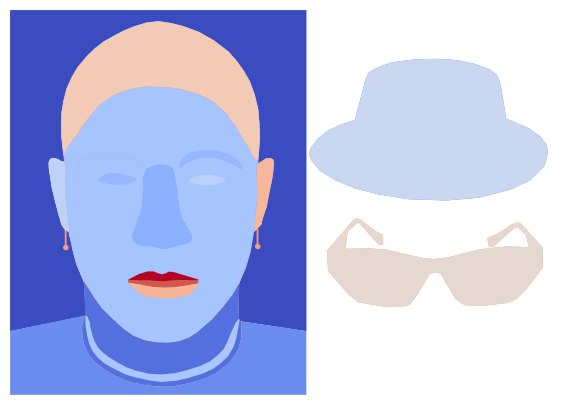}
        \caption{\footnotesize{Age - Lipstick}}
    \end{subfigure}
    \begin{subfigure}{0.16\linewidth}
        \centering
        \includegraphics[width=\linewidth]{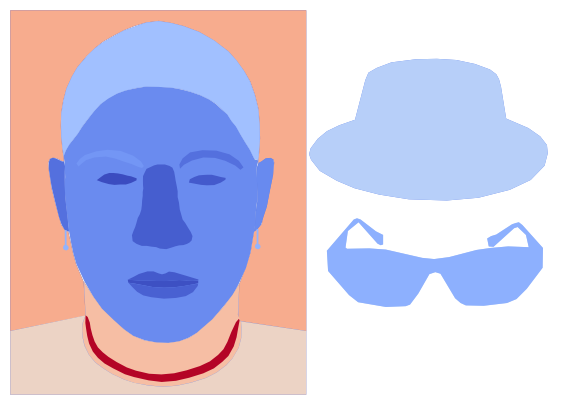}
        \caption{\footnotesize{Age - Necklace}}
    \end{subfigure}
    \begin{subfigure}{0.16\linewidth}
        \centering
        \includegraphics[width=\linewidth]{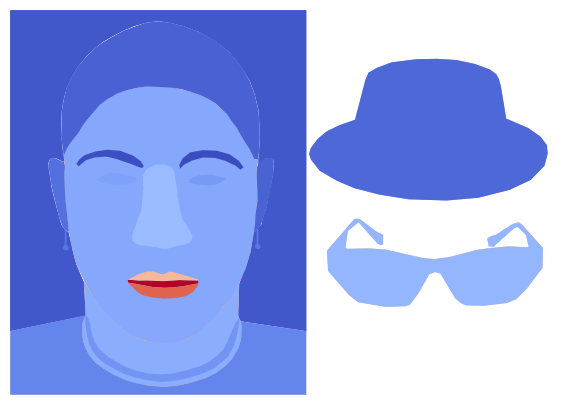}
        \caption{\footnotesize{Age - Smiling}}
    \end{subfigure}

    \caption{\METHOD\ heatmap output for single attribute bias experiments on CelebA (train) and CelebAMask-HQ (test). Heatmap's color scale is from blue to red, corresponding to the lowest and highest activations, respectively.}
    \label{fig:single_att}
\end{figure*}

\section{Results}
\subsection{Single attribute correlation}
In this set of experiments, we examine scenarios where a single facial attribute demonstrated a strong correlation, i.e., 99\%, with the target attributes - \texttt{Gender} and \texttt{Age}. 
Given that the chosen facial attributes, such as \texttt{Wearing\_Lipstick}, are inherently easier for models to learn than the actual targets (e.g., \texttt{Gender}), the models are expected to exploit these shortcuts. In such instances, \METHOD\ is anticipated to effectively identify biases stemming from these attribute-target correlations. Figure~\ref{fig:single_att} demonstrates the heatmap explanations provided by \METHOD , while Table~\ref{tab:single_att} presents the ranking position of the region connected to the biased attribute. As one may observe, \METHOD\ achieves a mean ranking position of 2.25 with a lowest possible rank of 6, which underscores \METHOD 's ability to consistently position the regions of interest within the top activated regions. 
\looseness=-1
\begin{table}[t]
\centering
\caption{
Evaluation of single attribute bias experiments on CelebA (train) and CelebAMask-HQ (test). FaceX's output ranking position for each target attribute w.r.t. the IoR values is reported.
}
\begin{tabular}{lcc}
\toprule
Target &  Attribute              & Ranking Position   \\
\midrule
Gender   & Blond\_Hair               & 3  \\
Gender   & Eyeglasses                & 6 \\
Gender   & Smiling                  & 1 \\
Gender   & Wearing\_Earrings       & 5 \\
Gender   & Wearing\_Lipstick         & 1 \\
Gender   & Wearing\_Necklace       & 2 \\
Age  & Blond\_Hair             & 2 \\
Age  & Eyeglasses               & 3 \\
Age  & Smiling                  & 1 \\
Age  & Wearing\_Earrings        & 1 \\
Age  & Wearing\_Lipstick         & 1 \\
Age  & Wearing\_Necklace      & 1 \\
\midrule
\multicolumn{2}{c}{Mean}          & 2.25 \\
\bottomrule
\end{tabular}
\label{tab:single_att}
\end{table}

Specifically, for the correlation between \texttt{Gender} and \texttt{Blond\_Hair}, the region of hair is placed 3rd. Here, note that lips and eyebrows are also in the top activated regions possibly due to the correlation between \texttt{Gender} and \texttt{Wearing\_Lipstick} attribute and the correlation between the hair color and the eyebrows color. In the case of \texttt{Age} target and \texttt{Blond\_Hair} attribute, the hair region is the second higher activated region. As regards the \texttt{Gender}-\texttt{Wearing\_Eyeglasses} experiment, the region of eyeglasses is placed 6th, suggesting that \METHOD\ struggles to effectively capture the presence of bias related to eyeglasses. Here, the top activated regions are eyes, eyebrows, and nose, implying that the model focuses on areas close to the eyeglasses or behind them, possibly due to limited training samples used for this experiment (i.e., 3114 positive training samples). 
For the \texttt{Age}-\texttt{Wearing\_Eyeglasses} experiment, the number of training samples depicting faces wearing eyeglasses is quite larger (i.e., 5322) and the eyeglasses are the 3rd higher activated region.
Regarding the experiments involving the \texttt{Smiling} attribute, the mouth is the top activated region for both \texttt{Age} and \texttt{Gender} targets. In this case, models demonstrate a clear focus on the region that introduces the bias. Table~\ref{fig:patches} illustrates the appearance of high-impact patches belonging to the regions of interest, where it is clear that the high-impact patches depict smiles. 
For \texttt{Gender}-\texttt{Wearing\_Earrings} case, like eyeglasses, the desired attribute is not placed in the top activated regions. The model's focus on the lips, as indicated by the top activated regions, suggests that the model is possibly affected by the correlation between \texttt{Wearing\_Earrings} and  \texttt{Wearing\_Lipstick} for the female class. This might be attributed to the model finding it easier to learn \texttt{Wearing\_Lipstick} than \texttt{Wearing\_Earrings}, resulting in lower activations for the latter. This is not the case for the \texttt{Age}-\texttt{Wearing\_Earrings} experiment, where earrings are the top activate region. 
In the case of \texttt{Wearing\_Lipstick} experiments, \METHOD\ performs notably well as lips is the top activated region for both \texttt{Gender} and \texttt{Age} targets, respectively. Note that \texttt{Wearing\_Lipstick} is connected to two regions (e.g., upper lip and lower lip), and thus \METHOD\ is expected to provide high activations for at least one of them. 
Finally, the region of the necklace is the second highest activated region for the \texttt{Gender}-\texttt{Wearing\_Necklace} experiment and the highest activated region for the \texttt{Age}-\texttt{Wearing\_Necklace} case. Here, both models demonstrate high activations in the region of interest.
\looseness=-1
\begin{table*}[h]
    \centering
        \caption{Appearance explanations for single attribute bias experiments on CelebA (train) and CelebAMask-HQ (test). The top 20 high-impact patches of the biased regions are reported.}
    \begin{tabular}{ccc}
        \toprule
        Target&Attribute&High-Impact Patches \\
        \midrule
        Age & Blond\_Hair & 
        \includegraphics[width=0.6\linewidth]{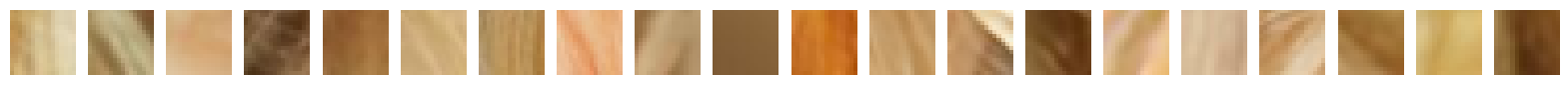} \\
        \midrule
        Age & Wearing\_Lipstick & 
        \includegraphics[width=0.6\linewidth]{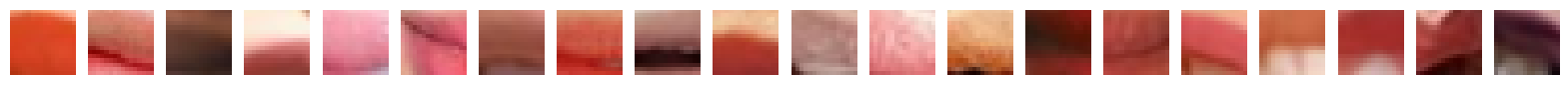} \\
\midrule
        Age & Smiling & 
        \includegraphics[width=0.6\linewidth]{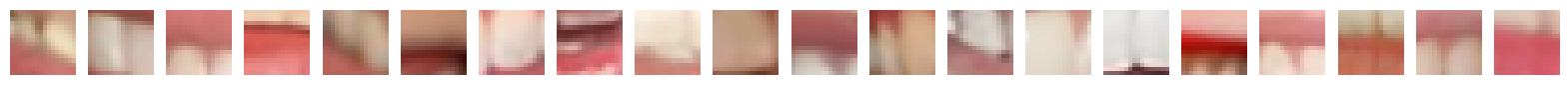} \\
\midrule
        Gender & Blond\_Hair & 
        \includegraphics[width=0.6\linewidth]{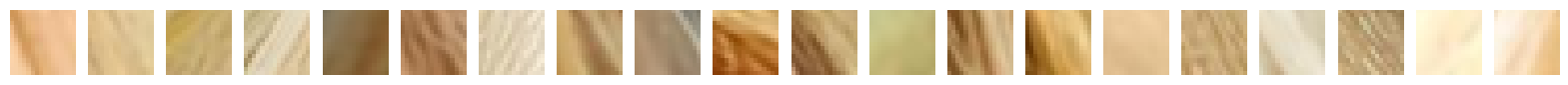} \\
        \midrule
        Gender & Wearing\_Lipstick & 
        \includegraphics[width=0.6\linewidth]{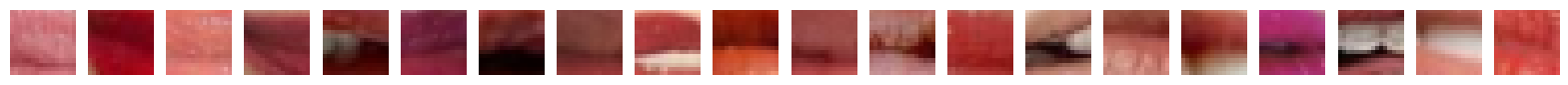} \\
\midrule
        Gender & Smiling & 
        \includegraphics[width=0.6\linewidth]{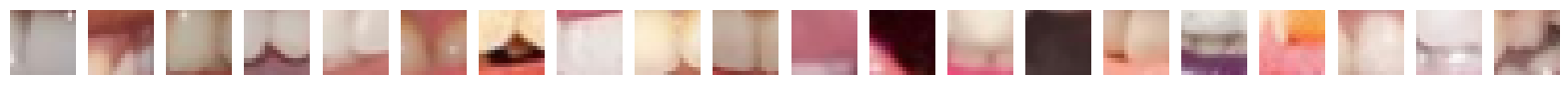} \\
        \bottomrule
    \end{tabular}
    \label{fig:patches}
\end{table*}
\subsection{Multiple attributes correlation}
In this subsection, we present experiments exploring correlations between multiple facial attributes and the target attributes - \texttt{Gender} and \texttt{Age}. The objective is to evaluate \METHOD 's performance in scenarios where biases arise from the combined influence of multiple attributes. 
\looseness=-1
\begin{table}[h]
\centering
\caption{Evaluation of multi-attribute bias experiments on CelebA (train) and CelebAMask-HQ (test). 
FaceX's output Ranking Position (RP) for each target attribute w.r.t. the IoR values is reported.
RP 1 and RP2 stand for the ranking position of the target region found first and second, respectively.}
\resizebox{\linewidth}{!}{
\begin{tabular}{lcccc}
\toprule
Target & Attribute 1 & Attribute 2     & RP 1   & RP 2  \\
\midrule
Gender   & Blond\_Hair & Smiling      & 1 & 11  \\
Gender   & Wearing\_Earrings & Smiling  & 1 & 11 \\
Gender   & Eyeglasses & Wearing\_Hat & 1 & 2  \\
Gender   & Wearing\_Earrings & Wearing\_Necklace  & 4 & 7 \\
Gender   & Wearing\_Lipstick & Wearing\_Necklace & 1 & 2 \\
Gender   & Smiling & Wearing\_Necklace  & 1 & 12 \\
Age  & Blond\_Hair & Smiling   & 1 & 10 \\
Age  & Wearing\_Earrings & Smiling   & 1 & 6 \\
Age  & Eyeglasses & Wearing\_Hat  & 1 & 4 \\
Age  & Wearing\_Earrings & Wearing\_Necklace  &  1 & 2 \\
Age  & Wearing\_Lipstick & Wearing\_Necklace  & 1 & 11 \\
Age  & Smiling & Wearing\_Necklace  & 1 & 2 \\
\midrule
\multicolumn{3}{c}{Mean}  & 1.25 & 6.67 \\
\bottomrule
\end{tabular}}
\label{tab:multi_att}
\end{table}
Similarly to the single attribute correlation, we assign a 99\% correlation between the target and two facial attributes. The results, summarized in Table~\ref{tab:multi_att}, \METHOD\ consistently positions at least one of the regions of interest within the top positions of the region ranking, with a mean ranking position of 1.25. However, only 33.33\% of the experiments position the regions related to both facial attributes to the top 3 regions, resulting in a mean ranking value of 6.67. 
This behavior can be attributed to inherent characteristics of deep learning models, where they tend to prioritize regions that can enhance their accuracy. In cases where focusing on one facial attribute alone is sufficient for accurate predictions, the model tends to do so. More accurately, the model's prioritization strategy depends on the relative difficulty of learning each attribute. In other words, a model is expected to focus on both facial attributes if they are both difficult to learn and on the easiest one, otherwise. 
Consequently, \METHOD\ often positions only one attribute in the top activated regions, a phenomenon expected in the context of deep learning models. Particularly, for \texttt{Gender} - \{\texttt{Eyeglasses} - \texttt{Wearing\_Hat}, \texttt{Wearing\_Lipstick} - \texttt{Wearing\_Necklace}\} and \texttt{Age} - \{\texttt{Eyeglasses} - \texttt{Wearing\_Hat}, \texttt{Wearing\_Earrings} - \texttt{Wearing\_Necklace}, \texttt{Smiling} - \texttt{Wearing\_Necklace}\} both facial attributes are placed in the top-ranked positions, while for the remaining experiments, models tend to rely on either first or the second facial attribute.

\subsection{Evaluation on RFW test benchmark}
\begin{figure}[h]
    \centering
    \begin{subfigure}{0.18\textwidth}
        \centering
        \includegraphics[width=\linewidth]{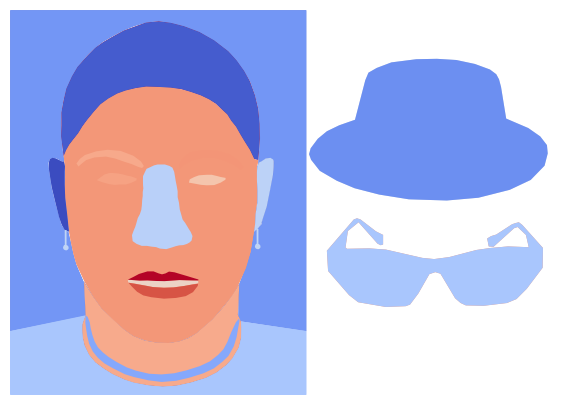}
        \caption{Heatmap}
        
    \end{subfigure}
    
    \begin{subfigure}{0.4\textwidth}
        \centering
        \includegraphics[width=\linewidth]{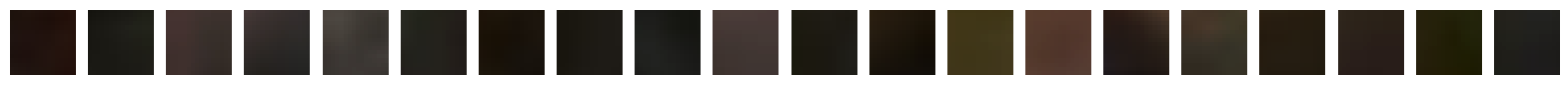}
        \caption{Skin high-impact patches}
        
    \end{subfigure}
    
    \caption{\METHOD\ heatmap output and high impact patches for single attribute bias experiment on FairFace (train) and RFW (test). Model's target is \texttt{gender} and the correlated attribute is \texttt{race}.  Heatmap's color scale is from blue to red, corresponding to the lowest and highest activations, respectively.}
    \label{fig:rfw}
\end{figure}

In this subsection, our attention shifts to the RFW test benchmark. Unlike CelebAMask-HQ, RFW lacks explicit information about the regions of interest. To predict these regions without explicit masks, FaRL is utilized (see Section~\ref{sec:farl}). 
The goal remains the exploration of biases, and to achieve this, we intentionally inject a strong correlation (99\%) between \texttt{Gender} and \texttt{Race} into the training data, i.e., FairFace, simulating a biased scenario. The objective is to assess whether \METHOD\ can discern and highlight regions associated with this introduced bias.
As presented in Figure~\ref{fig:rfw}, \METHOD\ successfully highlights the region of skin. Here the appearance of the skin region plays a crucial role in identifying the racial bias. As can be easily noticed, the high-impact patches associated with the identified region suggest a pronounced association with dark-skinned individuals.

\begin{figure}[t]
    \centering
    \begin{subfigure}{0.35\linewidth}
        \centering
        \includegraphics[width=\linewidth]{figures/single_att/male-blond-face.png}
        \caption{Vanilla}
        
    \end{subfigure}
    \begin{subfigure}{0.35\linewidth}
        \centering
        \includegraphics[width=\linewidth]{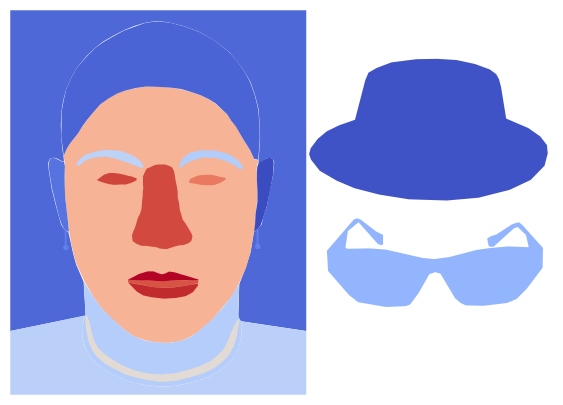}
        \caption{FLAC}
        
    \end{subfigure}
    
    \caption{\METHOD\ heatmap outputs for single attribute bias experiment on CelebA (train) and CelebAMask-HQ (test), before (i.e., Vanilla model) and after applying a bias mitigation approach (i.e., FLAC \cite{sarridis2023flac}). The model's target is \texttt{gender} and the correlated attribute is the \texttt{Blond\_Hair} attribute.  Heatmap's color scale is from blue to red, corresponding to the lowest and highest activations, respectively. }
    \label{fig:flac}
\end{figure}

\subsection{Impact of bias mitigation approaches}

Here, we explore the behavior of models on biased data with or without leveraging bias mitigation algorithms, simulating a scenario where a significant co-occurrence (99\%) exists between female individuals and \texttt{Blond\_Hair} attribute. 
The purpose of this experiment is to demonstrate the shift in FaceX's output distribution when applied to a fair model and a biased one.
Initially, the vanilla model tends to emphasize the hair and brow regions during predictions, aligning with the biased correlation introduced in the training data. To address and mitigate this bias, we employ FLAC \cite{sarridis2023flac}, a method that demonstrates state-of-the-art performance on training fair models using biased data. FLAC is introduced to counteract the influence of the biased correlation between \texttt{Gender} and \texttt{Blond\_Hair}, to foster fairer predictions.
The impact of FLAC becomes evident as there is a significant shift in the model's behavior, as presented in Figure~\ref{fig:flac}. Specifically, contrary to the initial training phase where hair and brows were among the top-activated regions, FLAC successfully mitigates this bias by significantly reducing the activations in the regions related to the \texttt{Blond\_Hair} attribute. It is worth noting, that further analyzing the FaceX outputs (e.g., inspecting the high-impact patches) for the FLAC-based model could reveal unknown biases that are ignored, however, this is beyond the scope of this paper. 
\looseness=-1
\subsection{Experiments on standard benchmarks}
In addition to the controlled scenarios, experiments are conducted on standard benchmarks without intentionally incorporating biases into the training data. Two models are trained, one on the CelebA dataset and another on FairFace, each serving as a representative of datasets with potentially biased and unbiased training data, respectively. The ground truth regions to recall are determined as the top 5 regions associated with facial attributes that most frequently co-occur with the target in the CelebA training data. These regions are computed based on the ground truth annotations.
The objective is to assess \METHOD 's effectiveness in identifying potential biases within the CelebA dataset when applied to a model trained on CelebA, where the presence of biases is indicated by high-ranking positions for the target regions. When \METHOD\ is applied to the model trained on FairFace, it is expected to observe low-ranking positions for the target regions (representing the CelebA biases), as FairFace does not exhibit the same biases found in CelebA.

\begin{figure}[t]
    \centering
    \begin{subfigure}{0.40\linewidth}
        \centering
        \includegraphics[width=\linewidth]{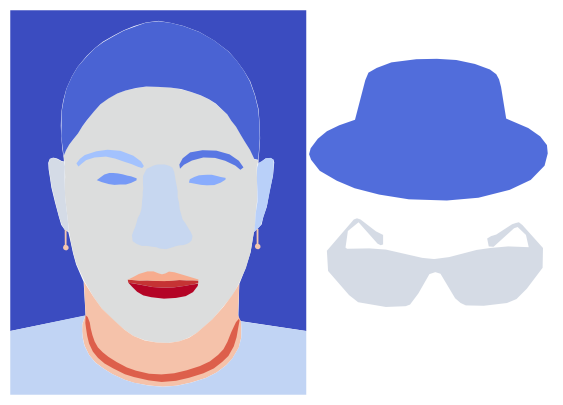}
        \caption{CelebA}
        
    \end{subfigure}
    \begin{subfigure}{0.40\linewidth}
        \centering
        \includegraphics[width=\linewidth]{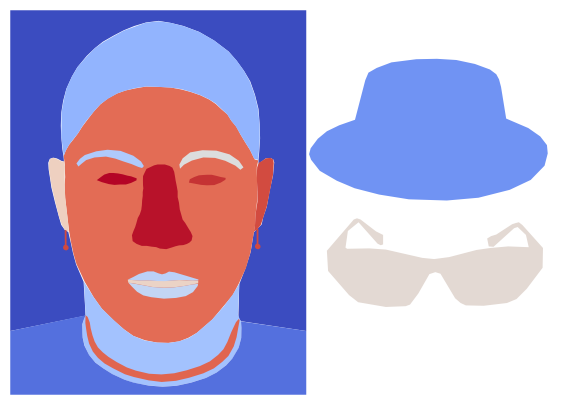}
        \caption{FairFace}
        
    \end{subfigure}
    
    \caption{\METHOD\ heatmap outputs for experiments on default on CelebA and FairFace training dataset with target the \texttt{Gender} attribute and CelebAMask-HQ as test benchmark. Heatmap's color scale is from blue to red, corresponding to the lowest and highest activations, respectively.}
    \label{fig:default}
\end{figure}

Table~\ref{tab:default} outlines the results of these experiments. Focusing on \texttt{Gender} prediction within the CelebA dataset, four out of five target regions are placed in the top 7 ranking positions, indicating that \METHOD\ effectively identifies four out of the five biased regions associated with \texttt{Gender} prediction.
In contrast, the evaluation of the FairFace model, acknowledged for its fairness concerning \texttt{Gender} and \texttt{Race}, demonstrates different behavior, with the first target region placed 4th in the ranking. This aligns with expectations, illustrating that the model trained on FairFace does not demonstrate the biases present in CelebA. However, the heatmap provided by \METHOD\ (see Figure~\ref{fig:default}) allows for identifying other potential biases involved in FairFace data. In particular, the model shows high activations on the regions of earrings, necklaces, and eyeglasses, which ideally should not be determinant features for \texttt{Gender} classification.
\begin{table}[t]
\centering
\caption{Evaluation on CelebA and FairFace training dataset with CelebAMask-HQ as test benchmark. The target regions are defined as the regions of the top 5 correlated attributes in CelebA, i.e., skin, lips, earrings, hair, and necklace. The RP values (RP 1-5) represent the ranking positions of the target regions' IoR values in ascending order. Notably, for CelebA, higher RP values suggest that the model is reproducing the biases inherent in CelebA. In contrast, for FairFace, lower RP values indicate that the model is less inclined to exhibit the biases observed in CelebA.}

\resizebox{\linewidth}{!}{
\begin{tabular}{lccccccc}
\toprule
Dataset &  Target                  & RP 1   & RP 2    & RP 3    & RP 4    & RP 5   \\
\midrule
CelebA   & Gender            & 1 (lips)& 3 (necklace)& 6 (earrings) & 7 (skin)  & 18 (hair) \\
FairFace   & Gender          & 4 (earrings)& 6 (necklace)& 7 (skin)& 13 (lips)& 16 (hair)\\
\bottomrule
\end{tabular}}
\label{tab:default}
\end{table}
\section{Conclusion}
In summary, \METHOD , our XAI methodology for facial analysis, addresses the critical need for comprehensive insights into model decisions. By providing spatial explanations across 19 facial regions and introducing appearance-oriented explanations through high-impact image patches, \METHOD\ goes beyond individual explanations. Through extensive evaluation in various controlled experiments, including scenarios with intentional biases and mitigation efforts, \METHOD\ demonstrates robustness and adaptability. Its application to real-world benchmarks, such as RFW, underlines its practical utility where explicit facial regions information may be lacking. In contrast to existing XAI approaches, \METHOD 's holistic view of model behavior contributes to the ongoing quest for fair and unbiased facial analysis systems. Future work could extend its application to other computer vision domains characterized by geometrical abstraction, fostering the development of fairer and more robust AI systems. Finally, potential research directions could focus on developing fairness-aware approaches that leverage FaceX explanations to mitigate biases during the training stage.
\begin{acks}
    This research was supported by the EU Horizon Europe project MAMMOth (Grant Agreement 101070285).
\end{acks}

\bibliographystyle{ACM-Reference-Format}
\balance
\bibliography{facex_arxiv}

\end{document}